\documentclass[letterpaper, 10 pt, conference]{ieeeconf}

\IEEEoverridecommandlockouts
\usepackage[T1]{fontenc}
\usepackage[utf8]{inputenc}
\usepackage{amsmath}
\usepackage{amssymb}
\usepackage{booktabs}
\usepackage{graphicx}
\usepackage{xcolor}
\usepackage{cite}
\usepackage{url}
\usepackage{balance}

\newcommand{\model}{\mbox{FLUX-mimic}}
\newcommand{\pifive}{\ensuremath{\pi_{0.5}}}

\title{\LARGE \bf Video-Action Models for\\Data-Efficient Industrial Manipulation}
\author{\parbox{0.98\textwidth}{\centering
Eugenio Chisari$^{1,*}$, Stefanos Charalambous$^{1,*}$, Oliver Sanchez$^{1,*}$, Thomas Umbach$^{1}$, Michael Gentner$^{2}$, Fabio Quattrini$^{1}$, Norica Bacuieti$^{1}$, Sabin Grube Doiz$^{2}$, Valentin Bertle$^{2}$, Sigmund Hennum Hoeg$^{1}$, Alain Ryser$^{1}$, German Rodriguez$^{1}$, Mario Arduini$^{1}$, Dang Nguyen$^{1}$, Florian Aumann$^{1}$, Felix Hegg$^{1}$, Manuel Meier$^{1}$, Philipp Wand$^{1}$, Julian Lotzer$^{1}$, Alessio Desogus$^{1}$, Benedek Forrai$^{1}$, Michael Bloesch$^{1}$, Stephan-Daniel Gravert$^{1}$, Stefan Weirich$^{1}$, Patrick Lanners$^{2}$, Elvis Nava$^{1}$\\[0.5em]
{\small $^{1}$Mimic Robotics\qquad $^{2}$BMW Group}}
  \thanks{$^{*}$Equal contribution.}
}

\IEEEaftertitletext{\makeatletter
\begin{minipage}{\textwidth}
    \def\@captype{figure}
    \centering
    \includegraphics[
        width=1.0\textwidth,
        keepaspectratio
    ]{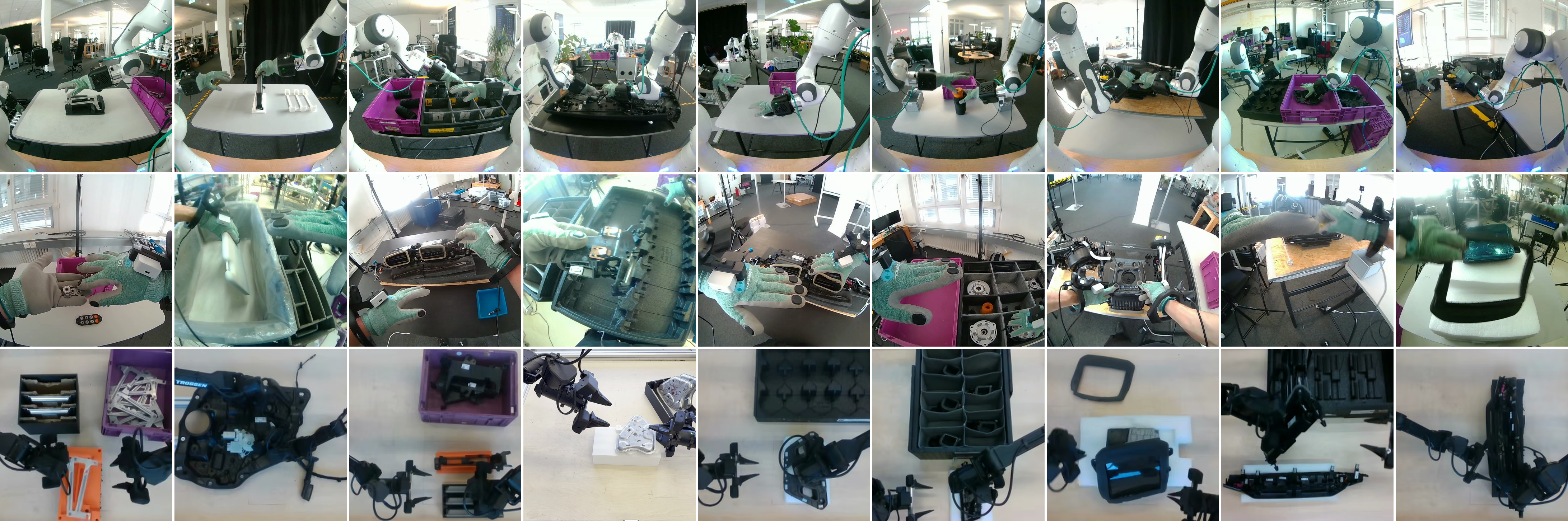}
    \caption{Industrial manipulation tasks from our automotive data collection, shown across the three collection sources used for pretraining: teleoperation of the dexterous bimanual platform (top), wearable data collection with human operators (middle), and teleoperation of a Trossen setup (bottom).}
    \label{fig:task_collage}
\end{minipage}
\makeatother\vspace{\baselineskip}}

\begin{document}
\maketitle
\thispagestyle{empty}
\pagestyle{empty}

\begin{abstract}
Industrial manipulation requires precise alignment and controlled handling of rigid, deformable, and flexible components.
Yet, the benefits of pretrained visuomotor policies remain underexplored in settings where limited access to specialized hardware, components, and fixtures constrains demonstration collection and evaluation.
Understanding task-specific data requirements is important for assessing their scalability across diverse processes on the factory floor.
We investigate whether video pretraining can support reliable industrial manipulation with limited action data.
We propose \model{}, a video-action model, motivated by the potential of video pretraining to capture motion and interaction patterns that enable sample-efficient training.
\model{} is built on a backbone pretrained on large-scale web and industrial task data, and an action expert to generate robot trajectories.
We train our policy by joint video-action training.
Across four automotive manufacturing benchmark tasks and two bimanual robot platforms, \model{} outperforms all considered baselines on the dexterous robot platform and matches \pifive{}'s average success rate on the gripper-based robot platform.
On a deformable part kitting task, \model{} reaches 77.5\% success with as little as two hours of task-specific action-labeled demonstrations.
\end{abstract}

\section{Introduction}
\label{sec:introduction}

Physical foundation models have been shown to bridge web-scale pretraining and real-world deployment, allowing robots to acquire manipulation behaviors that are difficult to specify through hand-engineered rules and controllers.
However, their adoption in industrial automation remains limited, as high task specificity with characteristic visual and physical properties and scarce demonstration data present significant hurdles.
For example, in the automotive industry, typical tasks include inserting components into rigid fixtures with tight clearances, routing flexible cables through constrained geometry, and picking parts from unstructured bins into ordered slots (see Fig.~\ref{fig:task_collage}), yielding scenes with metal panels, foam and rubber components, wire harnesses, and densely populated trays.
These tasks require precise control under visual and geometric variation, and are substantially different from standard web-scale priors.
% The domain gap between industrial automation and priors acquired during web-scale pretraining highlights the importance of precise control under high visual and geometric variation.
% Therefore, we add industrial task data into our video pretraining to capture scene dynamics from large-scale data and then adapt it to industrial tasks with a limited number of application-specific robot demonstrations.

Popular large-scale robot datasets, such as Open X-Embodiment~\cite{o2024open} and DROID~\cite{khazatsky2024droid}, provide diverse robot demonstrations with everyday objects in household environments.
Such generic datasets are often used during pretraining, but components and interactions encountered in industrial settings are not well covered.
To address this, we collect demonstrations of industrial tasks with materials and objects used in real production environments.

While vision-language-action models (VLAs), such as RT-2~\cite{zitkovich2023rt}, OpenVLA~\cite{kim2024openvla}, and \pifive{}~\cite{intelligence2025pi_}, provide strong language-visual semantics, industrial tasks place particular demands on physical reasoning about geometry, contact, and object dynamics.
Video-action models (VAMs), such as mimic-video~\cite{pai2025mimic} and DreamZero~\cite{ye2026world} address this by predicting robot actions based on representations from a generative video model. 
These representations inherently capture how a scene changes over time and provide a strong foundation for physical reasoning.

In this work, we present \model{}~\cite{nava2026fluxmimic}, a VAM trained on both video and action data of industrial tasks, that builds on top of a large-scale pretrained video backbone.
We study whether this approach improves performance and how it compares with alternative policies when action-labeled demonstrations are scarce.

We evaluate our policies on two bimanual platforms: Franka Robotics 7-DoF arms with dexterous hands, and Trossen Widow X robotic arms with two-finger grippers.
The benchmark covers four tasks from automotive manufacturing operations on Franka and two of these tasks on the Trossen set-up.
\model{} achieves higher or on par success rates when compared to \pifive{} and single-task flow matching policies.
We also study sample efficiency and show that \model{} exceeds both baselines' full-data performance using only one quarter of the action data for policy training.

The main contributions of this work are:
\begin{itemize}
    \item \model{}, a video-action model for industrial manipulation that combines a large-scale pretrained video backbone with joint video-action policy training.
    \item A real-robot evaluation on two bimanual robot platforms with different end effectors, covering four automotive tasks on Franka and two on Trossen.
    \item A study of sample efficiency on a deformable part kitting task, using six training data budgets, and an ablation of the joint video-action prediction objective used in our model.
\end{itemize}

\section{Related Work}
\label{sec:related_work}

\begin{figure*}[ht]
    \centering
    \includegraphics[width=\textwidth]{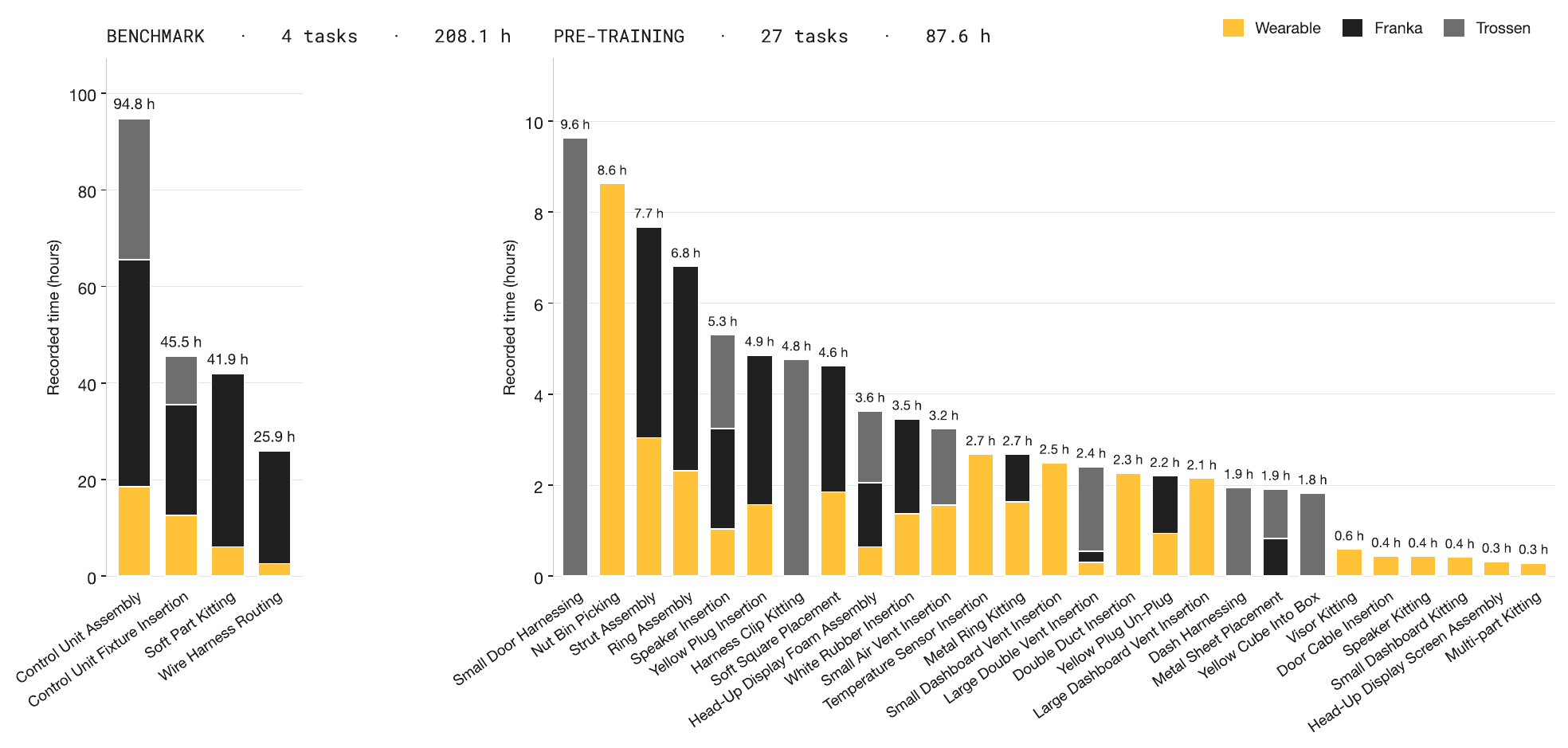}
    \caption{
        Recorded time per task in the industrial data
        collection, grouped into four designated benchmark
        tasks (left) and 27 additional pretraining tasks (right).
        The benchmark evaluation covers two robot platforms.
        Bars are stacked by collection source:
        wearable collection, Franka with dexterous hands teleoperation, and
        Trossen teleoperation. Both the benchmark and the 
        additional pretraining tasks are used for video-only backbone pretraining;
        benchmark-task recordings are additionally used for policy training and evaluation.
    }
    \label{fig:industrial_dataset_distribution}
\end{figure*}

Our work studies policy learning for industrial manipulation, with an emphasis on pretraining and the amount of action data required for policy training.

\subsection{Generative and Generalist Robot Policies}
Imitation learning has increasingly shifted toward expressive generative action representations.
Action Chunking with Transformers (ACT)~\cite{zhao2023learning} predicts sequences of actions to learn fine-grained bimanual manipulation, while Diffusion Policy~\cite{chi2023diffusion} and Robotic Diffusion Transformers~\cite{dasari2025dit} model complex multimodal action distributions through conditional denoising.
Flow matching~\cite{lipman2023flow} provides another continuous generative objective, used for action prediction in policies such as $\pi_0$~\cite{black2024pi0}.
Beyond task-specific architectures, generalist models leverage cross-task datasets: 
RT-1~\cite{brohan2022rt} and RoboCat~\cite{bousmalis2023robocat} investigate shared policies for multiple manipulation tasks, while Octo~\cite{octo2024} pretrains on Open X-Embodiment data and supports adaptation to new observation and action spaces.
VLAs additionally use pretrained visual and language representations: RT-2~\cite{zitkovich2023rt} transfers knowledge from vision-language training to control, while OpenVLA~\cite{kim2024openvla} studies generalist manipulation and adaptation.
\pifive{}~\cite{intelligence2025pi_} combines a pretrained vision-language backbone with a flow-matching action expert and learns from heterogeneous data spanning multiple robotic embodiments.
We use it as the main pretrained baseline to assess how a generalist VLA adapts to industrial manipulation compared with our video-pretrained policy.
Our benchmark also includes a task-specific flow matching policy.

\subsection{Pretraining and Data Efficiency}
Pretraining can supply visual representations or initialize a policy before learning a target task.
At the policy level, Octo~\cite{octo2024} and OpenVLA~\cite{kim2024openvla} demonstrate adaptation of generalist models using target-domain demonstrations.
These results motivate examining both the source of prior training experience and the amount of additional data required for a particular application.
Lin et al.~\cite{lin2024data} examine how demonstration quantity and environment diversity affect task-specific imitation learning.
They find diminishing benefits from additional demonstrations per object or environment, while broader coverage improves generalization.
At the multitask level, Barreiros et al.~\cite{barreiros2026careful} show that pretraining on diverse robot demonstrations improves downstream performance and reduces demonstration requirements for new tasks.
These findings motivate combining broad industrial task coverage for pretraining with concentrated demonstrations on selected benchmark operations.
For video-action models, Pai et al.~\cite{pai2025mimic} compare the sample efficiency of mimic-video and a $\pi_{0.5}$-style VLA by training action decoders on different-sized subsets of the simulated LIBERO~\cite{liu2023libero} benchmark, finding an advantage for video-conditioned action prediction.
Our study examines the same question on real-world industrial manipulation tasks, investigating the data requirements of different policy classes in this setting.

\subsection{Video Pretraining for Robot Control}
Video provides temporal supervision useful for robot learning.
Video Prediction Policy (VPP) ~\cite{hu2025vpp} uses predictive representations from a pretrained video diffusion model to guide robot action prediction.
mimic-video~\cite{pai2025mimic} uses intermediate representations from a pretrained video model to condition action prediction.
% : Latent Action Pretraining~\cite{ye2025lapa} learns latent actions from video, while DreamGen~\cite{jang2025dreamgen} generates visual trajectories for policy learning.
% Unified World Models~\cite{zhu2025uwm} couples video and action diffusion, providing a shared formulation for learning dynamics and control.
Other approaches jointly model visual futures and robot actions.
Cosmos Policy~\cite{kim2026cosmos} encodes actions, future observations, and values as latent frames within a video diffusion model, supporting both policy execution and planning.
LingBot-VA~\cite{li2026causal} uses an autoregressive mixture of transformers to couple visual prediction and action inference, incorporating observations from execution into its temporal context.
DreamZero~\cite{ye2026world} jointly models video and actions using a pretrained video backbone and heterogeneous robot data, studying generalization to new tasks and adaptation to new embodiments.

Video supervision during training can also benefit policies that do not generate future videos at deployment.
Fast-WAM~\cite{yuan2026fastwam} retains video co-training but skips future prediction at inference; its ablations find a larger performance drop from removing video co-training than from omitting test-time video generation.
\model{} is trained on the joint video-action prediction objective, but does not require a full rollout of the video modality at inference time. 

\subsection{Industrial Manipulation and Data Collection}
Large robot datasets support learning across tasks and environments, but differ in their embodiment coverage, sensory information, and collection procedures.
Open X-Embodiment~\cite{o2024open} brings together demonstrations from multiple robotic platforms, while DROID~\cite{khazatsky2024droid} emphasizes collection across diverse real-world scenes.
AgiBot World~\cite{bu2025agibot} further expands the scale of manipulation data collection.
RH20T~\cite{fang2023rh20t} includes contact-rich skills with synchronized visual, force, audio, and action information.
For assembly, FMB~\cite{luo2024fmb} uses reproducible 3D-printed components and sequences of grasping, reorientation, and insertion, enabling evaluation across object geometries and multistage tasks.
Kernbach et al.~\cite{kernbach2026behavioral} study behavioral cloning for connector insertion using visual and force-torque observations, addressing variation in connector geometry and pose.
These works provide relevant settings for learning precise interactions and assembling parts.
Our data collection focuses on automotive manufacturing operations identified through observations of manual production work.
It includes production components and fixtures, with tasks spanning rigid-part assembly, insertion, cable routing, and kitting of deformable parts.

\section{Industrial Data Collection}
\label{sec:data}

\subsection{Task Identification and Selection}
Examining automotive manufacturing sites, we identified approximately 40 candidate manipulation tasks among operations performed manually.
We then filtered these tasks based on various selection criteria, which included levels of dexterity, payload constraints, and workspace reach of the robotic platforms.
Collected tasks were then organized into five categories according to their manipulation requirements, as described below.

\subsection{Manipulation Categories}
\paragraph{Cabling}
These tasks require routing flexible objects, such as wire harnesses, through rigid geometry and fixing them at several locations along their length. 
The robot must account for the cable's changing shape while positioning it relative to the surrounding structure.

\paragraph{Soft-Body Manipulation}
These tasks involve grasping and placing deformable components, such as foams and gaskets. 
Their shape changes under contact, affecting both the grasp and the placement of the part.

\paragraph{Insertion}
These tasks require mating a rigid component with a rigid fixture. 
Tight clearances and constrained approach directions make alignment important, while variation in the initial part presentation adds a perception requirement.

\paragraph{Assembly}
These tasks join two or more rigid parts into a subassembly. 
Examples include mating parts through snap-fit or screw connections. 
The category captures operations in which the relative arrangement of multiple components must be controlled during contact.

\paragraph{Kitting}
These tasks require picking parts from an unstructured bin and placing them into designated slots. 
The robot must handle variation in the source arrangement and produce an ordered final configuration.  Kitting can also involve soft-body manipulation when the parts are deformable.

\subsection{Data Collection}
After identification, we collected data through both robot teleoperation and camera-only wearable recording.
Teleoperation used two bimanual platforms: Franka arms with dexterous hands and Trossen arms with grippers.
Both platforms are equipped with a static camera, plus two wrist-mounted cameras.
The wearable setup consists of gloves with wrist-mounted cameras and a static torso camera to mirror the robot setups.
It records videos of human task execution without action labels and is used during pretraining only.

The collection summarized in Fig.~\ref{fig:industrial_dataset_distribution} comprises 31 distinct tasks, with 55,217 recorded episodes totaling 295.8 hours. 
Robot teleoperation accounts for 40,098 episodes and 218.9 hours across 20 tasks.
Wearable collection contributes a further 15,119 episodes and 76.8 hours, including 11 tasks absent from the teleoperation collection. 
The distribution reflects extensive collection on four designated benchmark tasks and smaller collections across 27 additional tasks that broaden industrial coverage. 
We evaluate all four benchmark tasks on Franka and two benchmark tasks on Trossen, as described in Section~\ref{sec:benchmark_tasks}.

\section{Technical Approach}
\label{sec:method}
\model{} combines a video backbone pretrained on web and industrial data with an action expert for industrial manipulation.

\subsection{Video Pretraining}
\label{sec:video_pretraining}
We initialize the video backbone of \model{} from FLUX 3~\cite{bfl2026flux3}, a multimodal foundation model jointly pretrained on images, video, and audio. 
The FLUX 3 model is pretrained on millions of hours of general video content, including hundreds of thousands of hours of video depicting human and robot manipulation. 
Our entire industrial dataset, including the benchmark tasks, is part of the video backbone’s pretraining.
This broad collection provides visual experience across objects, environments, embodiments, and interactions relevant to industrial manipulation, enabling its latent representations to capture how objects move, deform, and interact throughout a sequence in diverse settings.

FLUX 3 is trained with Self-Flow~\cite{chefer2026self}, which combines a flow matching objective with self-supervised feature prediction.
Flow matching~\cite{lipman2023flow} learns a velocity field along paths between noise and data.
Self-Flow additionally trains early representations to predict features obtained from a less corrupted view of the same input, akin to a teacher-student paradigm.
Its dual-timestep scheduling samples two noise levels and assigns them to different subsets of input tokens using a random mask.
In self-flow, these tokens are passed to a student model, while a teacher receives an input with all tokens perturbed at the lower of the two noise levels.
The student is trained to both predict the flow target for its mixed-noise input, and to align early layer representations to the teacher's features.
The objective can be summarized as
\begin{equation}
    \mathcal{L}_{\mathrm{pretrain}} = \mathcal{L}_{\mathrm{video}}
    + \lambda_{\mathrm{rep}}\mathcal{L}_{\mathrm{rep}},
    \label{eq:video_pretraining}
\end{equation}
where $\mathcal{L}_{\mathrm{video}}$ is the flow prediction loss, $\mathcal{L}_{\mathrm{rep}}$ aligns the student and teacher features, and $\lambda_{\mathrm{rep}}$ weights the representation objective.
The feature targets for \model{} come from the teacher, which is the EMA of the video model itself, without a separately pretrained representation encoder~\cite{chefer2026self}.

This objective is relevant to our policy architecture because the action expert operates on intermediate backbone features, motivating our choice for a backbone trained to support both generation and representation learning.

\subsection{Video-Action Model Architecture}
\label{sec:vam_architecture}

\begin{figure}[t]
    \centering
    \includegraphics[width=1\columnwidth, trim={100 110 100 110}, clip]{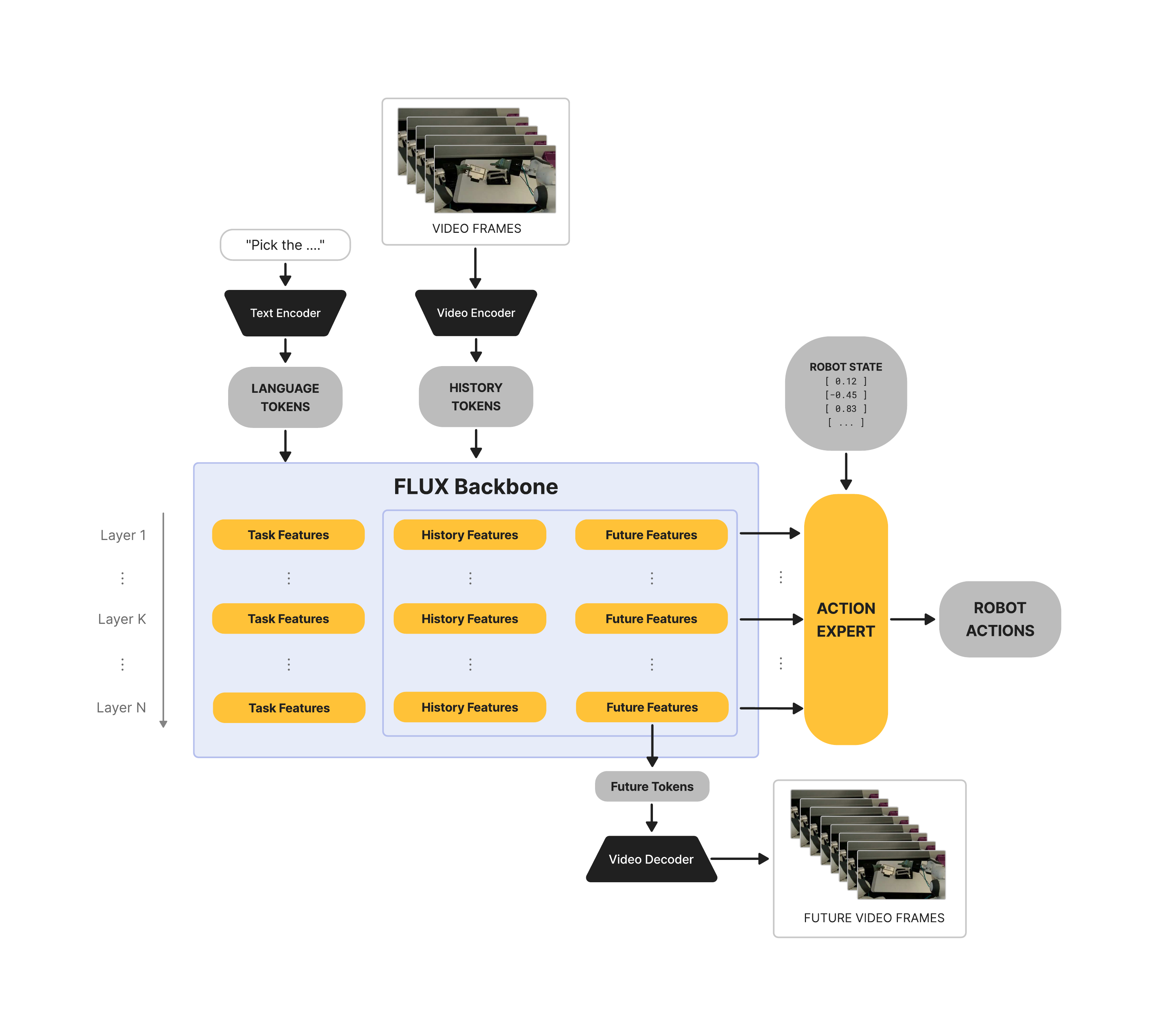}
    \caption{%
        \textbf{\model{} architecture}. We condition the video backbone with the task caption and the past video frames and the action expert with the past robot state. Then, we run one video denoising step and layer-wise attend to all video features with our action expert to predict robot actions. We train with joint video-action loss.
    }
    \label{fig:I-VAM-Architecture}
\end{figure}

\model{} combines the pretrained video backbone with a compact action expert that predicts robot action sequences, as illustrated in Figure~\ref{fig:I-VAM-Architecture}.
The video backbone receives video observations and is conditioned on the task language instruction.
Its hidden features represent the predicted evolution of the manipulation scene. 
The action expert receives robot states alongside the backbone features to predict robot actions and has the same depth as the video backbone, with each expert layer attending to the features from the corresponding backbone layer.
This conditions action prediction on the backbone’s representation of how the scene is expected to evolve.

At inference time, one backbone denoising step provides the visual features used by the action expert.
The expert then denoises an action chunk conditioned on these features, without requiring a complete video sampling procedure.

The backbone retains its video prediction capability after adding the action decoder.
This allows us to supervise both video and action prediction during policy training.

\subsection{Policy Training}
\label{sec:policy_training}
Policy training adapts the pretrained video representation to industrial manipulation through joint video-action prediction.
We train both modalities with conditional flow matching~\cite{lipman2023flow}, using demonstrations to supervise the scene's evolution and the corresponding robot actions.
For modality $m\in\{\mathrm{action},\mathrm{video}\}$, let $x^m$ denote the clean training target: an action chunk or a sequence of video latents.
We construct a noisy input by interpolating between Gaussian noise $\epsilon^m\sim\mathcal{N}(0,I)$ and the target according to a timestep $t_m$,
\begin{equation}
    x^m_{t_m} = (1-t_m)x^m + t_m\epsilon^m.
    \label{eq:policy_flow_path}
\end{equation}
Throughout this work, $t_m=0$ corresponds to clean data and $t_m=1$ to pure noise.
The target velocity along this path is $u^m=\epsilon^m-x^m$. Action sampling integrates the learned velocity field from $t_{\mathrm{action}}=1$ to $t_{\mathrm{action}}=0$.
The modality-specific objective is

\begin{equation}
    \mathcal{L}_m = \mathbb{E}_{x^m,\;\epsilon^m,\;t_m \sim p_m}\!\left[
        \operatorname{MSE}\!\left(\hat{u}^m, u^m\right)
    \right],
    \label{eq:policy_flow_loss}
\end{equation}
where $\hat{u}^m$ is the predicted velocity and the expectation is over demonstrations, noise, and training timesteps drawn from the modality-specific schedule $p_m$, specified below.

Interpolation and velocity loss are defined in the video model's latent space for video and in normalized action space for actions. We embed state and actions with separate single linear projection layers.

For video, we set $p_m$ following the Plateau-Logit-Normal schedule from~\cite{chefer2026self}. With $p_{\mathrm{ln}}(t)$ being the Logit-Normal distribution, this can be defined as
% \begin{equation}
%     p_{\mathrm{pln}}(t) \;\propto\; p_{\mathrm{ln}}\!\big(\min(t, t^\star)\big),
%     \qquad t^\star = \arg\max_t\, p_{\mathrm{ln}}(t).
% \end{equation}

\begin{equation}
    p_{\mathrm{pln}}(t) \propto
    \begin{cases}
        p_{\mathrm{ln}}(t), & t \le t^\star, \\[2pt]
        p_{\mathrm{ln}}(t^\star), & t > t^\star,
    \end{cases}
    \quad t^\star = \arg\max_t\, p_{\mathrm{ln}}(t).
\end{equation}

For actions, we set $p_m$ to the beta timestep sampling schedule of \pifive{}~\cite{intelligence2025pi_}. This can be written as
\begin{equation}
    b\sim\operatorname{Beta}(1.5,1),\quad
    t_{\mathrm{action}}=s b+(1-s),\quad s=0.999.
    \label{eq:policy_timestep_sampling}
\end{equation}
These distributions place more weight on noisier inputs.
During training, we draw timestep samples from the two modality schedulers independently.

We combine the two flow matching objectives as
\begin{equation}
    \mathcal{L}_{\mathrm{policy}} = \mathcal{L}_{\mathrm{action}}
    + \lambda_{\mathrm{video}}\mathcal{L}_{\mathrm{video}},
    \label{eq:policy_training}
\end{equation}
with $\lambda_{\mathrm{video}}=1$ in all runs, giving the two loss terms equal weight.
The action objective trains the action expert to predict demonstrated behavior from the backbone features and robot state, while the video objective retains supervision for visual prediction during policy training.

\section{Experimental Evaluation}
\label{sec:experiments}

We evaluate \model{} on two bimanual robot platforms at different locations, using the automotive tasks described in Section~\ref{sec:benchmark_tasks}.
We investigate how our method compares to baseline policies on each platform and how performance changes when less task data is available.
We also examine the contribution of the video prediction objective during policy training.

\subsection{Benchmark Tasks and Setup}
\label{sec:benchmark_tasks}
\begin{figure}[t]
    \centering
    {\footnotesize\textbf{Franka with dexterous hands}\par}
    \smallskip
    \begin{minipage}[t]{0.485\columnwidth}
        \centering
        \includegraphics[
            width=\linewidth,
            trim={0bp 60.85bp 0bp 60.85bp},
            clip
        ]{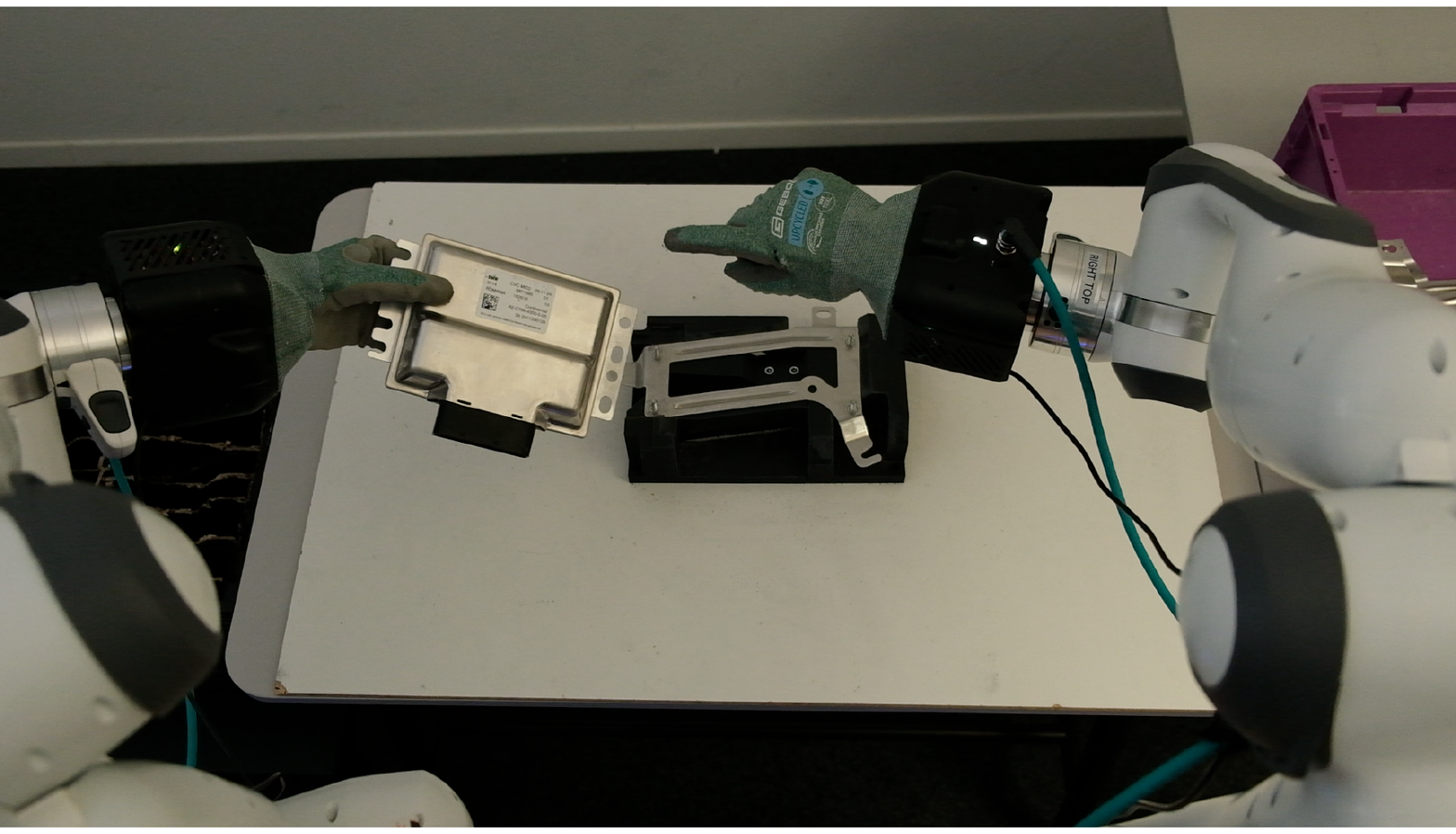}
        {\footnotesize (a) Control unit assembly}
    \end{minipage}
    \hfill
    \begin{minipage}[t]{0.485\columnwidth}
        \centering
        \includegraphics[
            width=\linewidth,
            trim={0bp 60.85bp 0bp 60.85bp},
            clip
        ]{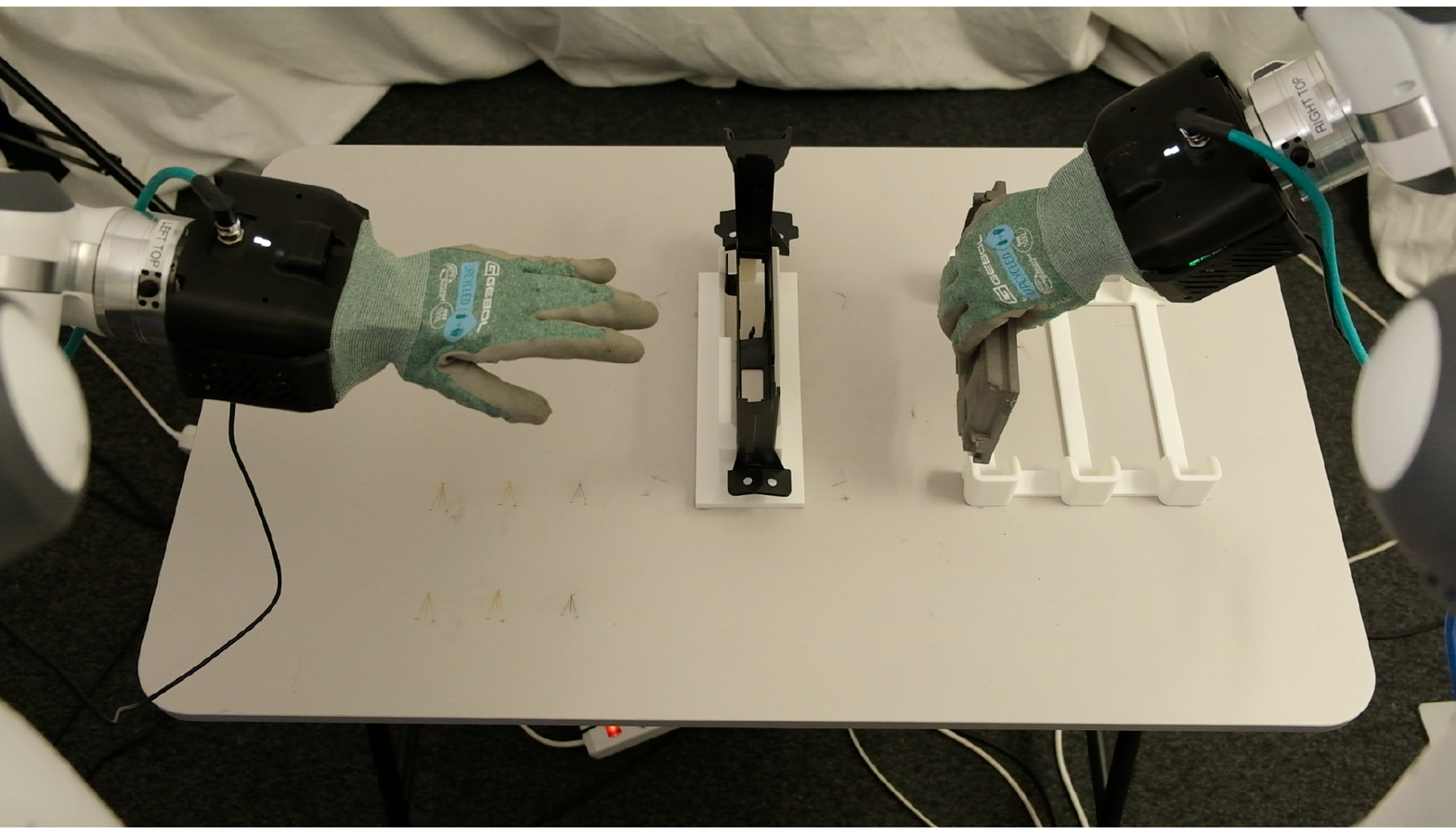}
        {\footnotesize (b) Control unit fixture insertion}
    \end{minipage}

    \par\medskip

    \begin{minipage}[t]{0.485\columnwidth}
        \centering
        \includegraphics[
            width=\linewidth,
            trim={0bp 60.85bp 0bp 60.85bp},
            clip
        ]{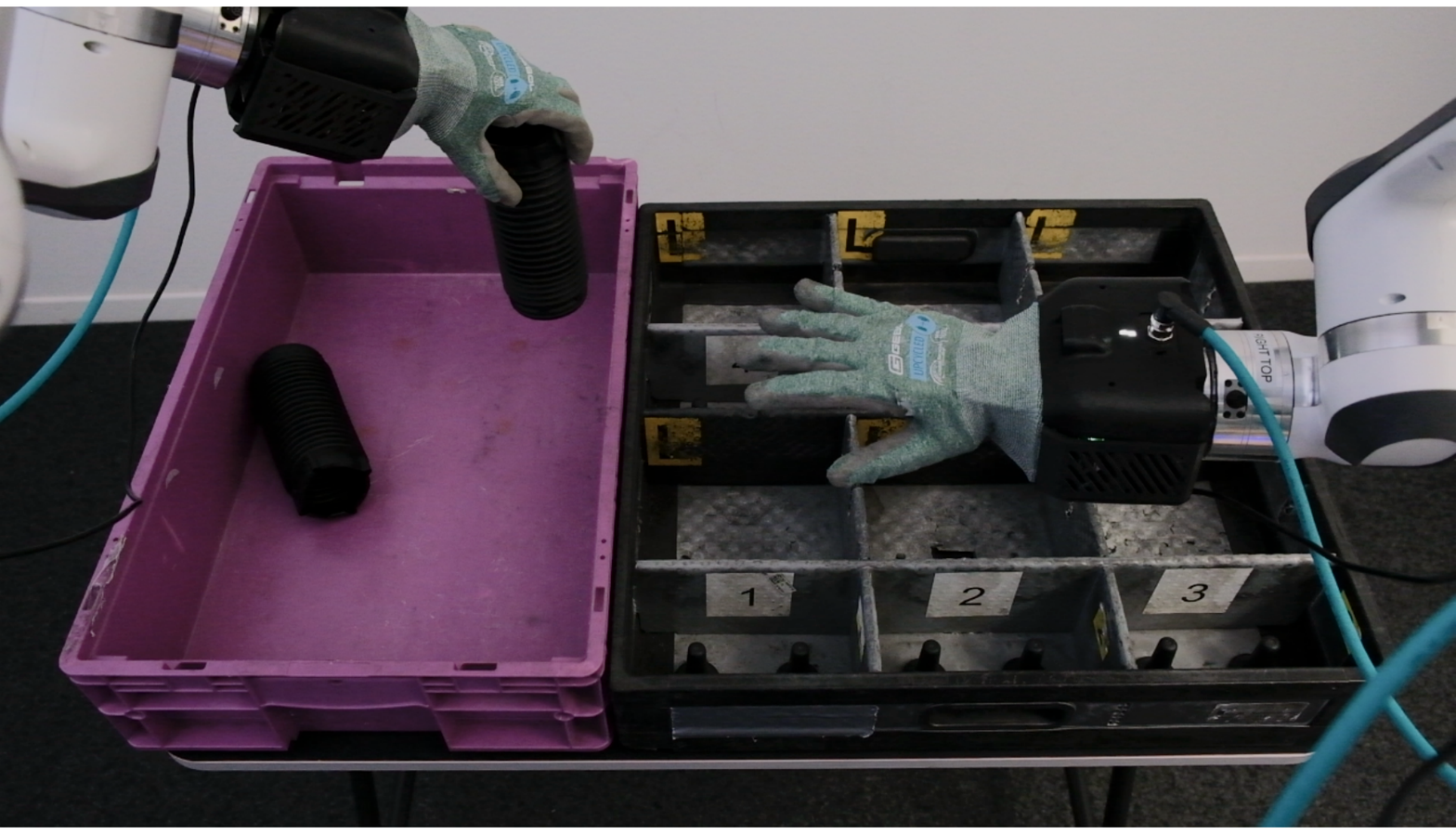}
        {\footnotesize (c) Soft part kitting}
    \end{minipage}
    \hfill
    \begin{minipage}[t]{0.485\columnwidth}
        \centering
        \includegraphics[
            width=\linewidth,
            trim={0bp 60.85bp 0bp 60.85bp},
            clip
        ]{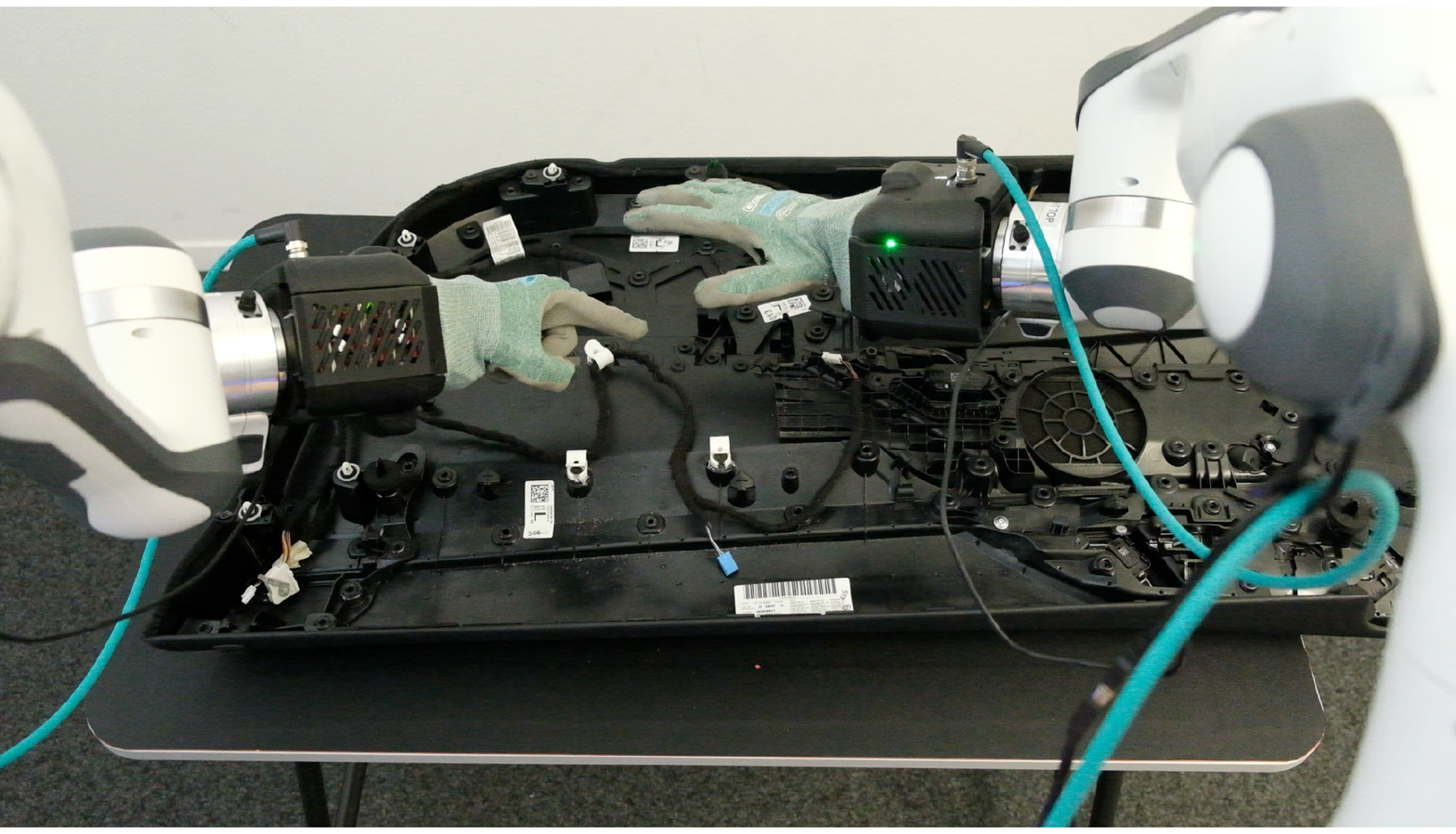}
        {\footnotesize (d) Wire harness routing}
    \end{minipage}

    \par\medskip
    {\footnotesize\textbf{Trossen with grippers}\par}
    \smallskip
    \begin{minipage}[t]{0.485\columnwidth}
    \centering
    \includegraphics[
        width=\linewidth,
    ]{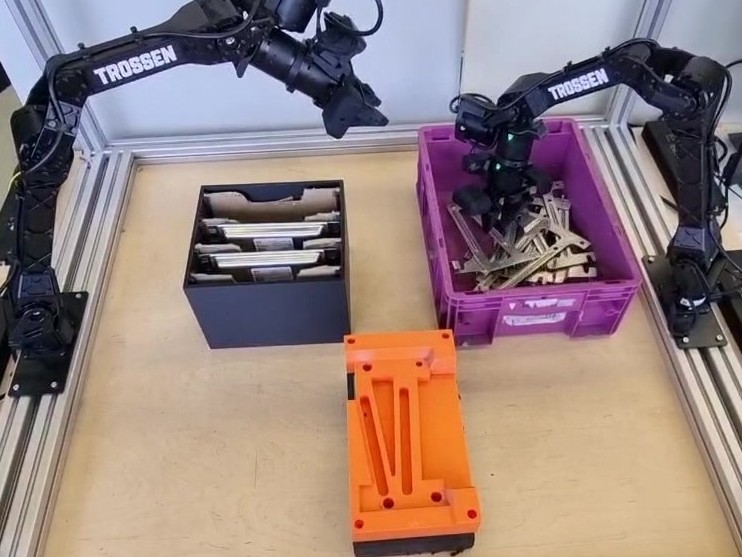}
    {\footnotesize (e) Control unit assembly}
    \end{minipage}
    \hfill
    \begin{minipage}[t]{0.485\columnwidth}
        \centering
        \includegraphics[
            width=\linewidth,
        ]{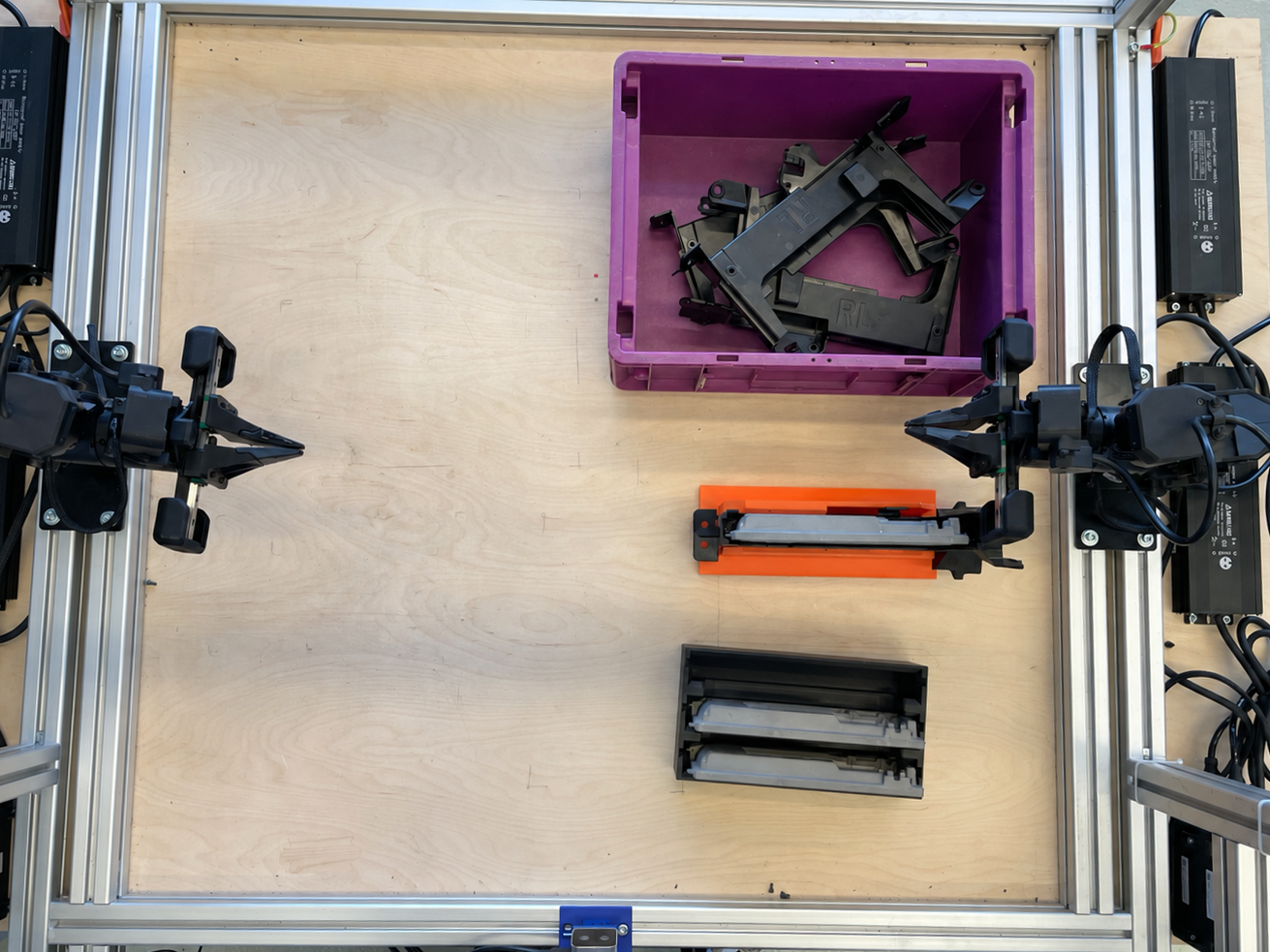}
        {\footnotesize (f) Control unit fixture insertion}
    \end{minipage}

    \caption{
        Industrial manipulation task setups.
        \textbf{Franka with dexterous hands (a--d):}
        control unit assembly, control unit fixture insertion, soft part kitting, and wire harness routing.
        \textbf{Trossen with grippers (e--f):}
        control unit assembly and control unit fixture insertion.
    }
    \label{fig:benchmark_tasks}
\end{figure}

We use two bimanual platforms: Franka arms with dexterous hands and Trossen arms with grippers.
Both platforms use the camera arrangement described in Section~\ref{sec:data}.
The platforms operate at different locations, and policies are trained separately for each embodiment.
To ensure reproducibility, our evaluation experiments use a dedicated cell with uniform lighting and background.
Figure~\ref{fig:benchmark_tasks} shows representative setups for the benchmark tasks across the two embodiments.

The Franka evaluation covers control unit assembly, control unit fixture insertion, soft part kitting, and wire harness routing.
The Trossen evaluation includes control unit assembly and control unit fixture insertion.
Together, these tasks cover deformable parts, assembly, insertion, kitting, and cabling.

\paragraph{Control unit assembly}
The robot retrieves a control unit from a box and mounts it onto an aluminium bracket. This task requires grasping the unit and accurately aligning it with the bracket during mounting.

\paragraph{Control unit fixture insertion}
The robot inserts a control unit component into a rigid fixture. The task requires aligning the component with the fixture and maintaining this alignment during insertion.

\paragraph{Soft part kitting}
The robot picks deformable foam strut covers from a bin and places them into ordered slots.
The task combines grasping from an unstructured source arrangement with placement of soft parts.
We also use this task for the sample-efficiency experiment and the video loss ablation.

\paragraph{Wire harness routing}
The robot routes a flexible cable along a door panel and fixes it at the required locations.
Successful execution requires handling the cable while positioning it relative to the panel geometry.

\subsection{Evaluation Protocol}
\label{sec:evaluation_protocol}
For each evaluated checkpoint, we report the fraction of rollouts that complete the full task as the task success rate.
We count a rollout as successful when the policy completes the task according to its terminal success criterion.
If the robot remains stuck on a stage for 30 seconds, we count the rollout as a failure and stop execution.

For each task, the evaluation protocol defines discrete task variations and distributes rollouts evenly across them.
These variations include different object orientation and position in the environment.
Control unit assembly, control unit fixture insertion, and soft part kitting each have five variations, while wire harness routing has six.
Rollout budgets are chosen as multiples of the number of variations to ensure equal coverage within each task.
We use 20 rollouts per method for assembly, fixture insertion, and kitting (four per variation), and 18 for wire harness routing (three per variation).
Under the controlled setup described in Section~\ref{sec:benchmark_tasks}, task configuration is the intended source of variation in the comparison.

\subsection{Benchmark Results}
\label{sec:benchmark_results}

\begin{table}[t]
    \caption{Full-task success rates (\%) for Franka and Trossen policies.}
    \label{tab:benchmark_results}
    \centering
    \small
    \setlength{\tabcolsep}{1.5pt}
    \begin{tabular*}{\columnwidth}{@{\extracolsep{\fill}}lccccc@{}}
        \toprule
        Method & \shortstack{Control\\unit\\assembly} & \shortstack{Control unit\\fixture\\insertion} & \shortstack{Soft part\\kitting} & \shortstack{Wire\\harness\\routing} & Avg. \\
        \midrule
        \multicolumn{6}{c}{\textit{Franka with dexterous hands}} \\
        \midrule
        FM & 50 & 35 & 70 & 55.6 & 52.6 \\
        \pifive{} & 0 & 30 & 75 & 5.6 & 27.6 \\
        \model{} (ours) & \textbf{85} & \textbf{85} & \textbf{100} & \textbf{61.1} & \textbf{82.8} \\
        \midrule
        \multicolumn{6}{c}{\textit{Trossen with grippers}} \\
        \midrule
        \pifive{} & \textbf{60} & 60 & - & - & \textbf{60.0} \\
        \model{} (ours) & 50 & \textbf{70} & - & - & \textbf{60.0} \\
        \bottomrule
    \end{tabular*}
\end{table}

We compare the following methods:
\begin{itemize}
    \item \textbf{Single-task FM}: a policy with a pretrained CLIP ViT-B/16 image encoder and a transformer action predictor inspired by the DiT-Block architecture of Dasari et al.~\cite{dasari2025dit}. We modify the generative objective to use flow matching~\cite{lipman2023flow}, training the action predictor with an MSE loss on flow velocities.
    \item \textbf{\pifive{}}~\cite{intelligence2025pi_}: the VLA baseline finetuned on the benchmark tasks using the OpenPI implementation\footnote{\url{https://github.com/Physical-Intelligence/openpi}}. We fine-tune the released generalist checkpoint, whose reported pretraining data mix does not include industrial task data. For the Franka platform with dexterous hands, we initialize a new action projection to accommodate its larger action space.
    \item \textbf{\model{} (ours)}: the VAM trained on industrial task data via joint video-action prediction. The video backbone is pretrained on web-scale data as well as industrial manipulation video. The video-action policy is then trained on our industrial pretraining dataset.
\end{itemize}

We train separate \model{} and \pifive{} policies for Franka and Trossen.
For all benchmark tasks, \model{} and \pifive{} are conditioned on the task-specific language instructions listed in Table~\ref{tab:language_instructions}. 
Each \model{} and \pifive{} policy is trained jointly across tasks, while a separate FM policy is trained for each task.

We train all policies until convergence using AdamW.
For FM, we use a learning rate of $3\times10^{-4}$, a weight decay of $10^{-6}$, a cosine learning rate schedule and a batch size of 384.
For \pifive{}, we follow the OpenPI training recipe with a learning rate of $2.5\times10^{-5}$, a weight decay of $10^{-10}$, a batch size of 64, and 1,000 linear warmup steps followed by a constant learning rate.

For \model{}, we use separate optimizer parameter groups for the action expert and the pretrained video backbone.
The action expert uses a learning rate of $3\times10^{-4}$ and a weight decay of $0.1$, while the backbone uses a lower learning rate of $10^{-4}$ and no weight decay.
We use 1,000 steps of linear warmup followed by a constant learning rate schedule and a batch size of 256. 
For video prediction, we use 5 conditioning frames and 8 future frames at 5 FPS. 
Actions are predicted at 15 Hz with a horizon of 20.

\begin{table}[t]
\caption{Language instructions used to condition \model{} and \pifive{}.}
\label{tab:language_instructions}
\centering
\footnotesize
\begin{tabular}{p{0.30\linewidth}p{\dimexpr0.70\linewidth-4\tabcolsep\relax}}
\toprule
Task & Language instruction \\
\midrule
Control unit assembly &
Place the metal piece on the jig and place the control unit component on top of it. \\
Control unit fixture insertion &
Pick the black fixture, place it on the white jig, pick the metal control unit, and insert it onto the black fixture. \\
Soft part kitting &
Pick a black strut cover from the purple bin and fill the middle narrow row from left to right, with the yellow part facing downward. \\
Wire harness routing &
Clip the black cable into the three white clips on the black panel. \\
\bottomrule
\end{tabular}
\end{table}

At deployment, we execute the first 8 actions from each chunk.  Commands are executed at 30 Hz on Franka and 100 Hz on Trossen. Franka uses task-space actions and nearest-neighbor selection for temporal action ensembling, whereas Trossen uses joint-space actions and linear interpolation. All policies are trained and deployed with real-time chunking (RTC) ~\cite{black2025trainingtime}.

Table~\ref{tab:benchmark_results} reports full-task success rates for both platforms.
We compute averages separately over the tasks evaluated on each platform.

On Franka, \model{} achieves success rates of 85\%, 85\%, 100\%, and 61.1\% on control unit assembly, control unit fixture insertion, soft part kitting, and wire harness routing, respectively.
It outperforms both baselines on all four tasks, with an average success rate of 82.8\%, compared to 52.6\% for FM and 27.6\% for \pifive{}.
The advantage over FM is largest on control unit fixture insertion (50 percentage points), followed by assembly (35 points) and kitting (30 points), and is smaller on wire harness routing (5.6 points).
On control unit assembly, \pifive{} completes the initial steps of placing the metal piece and picking up the control unit, but fails to insert the control unit precisely, resulting in no full-task successes.
This illustrates how successful execution of early stages can still fall short of completing a task that requires precise insertion.

On Trossen, \model{} achieves 70\% success on control unit fixture insertion, compared to 60\% for \pifive{}.
\model{} achieves 50\% on control unit assembly, compared to 60\% for \pifive{}.
Both methods achieve an average success rate of 60.0\% across the two Trossen tasks.

The relative performance of the two methods varies across platforms.
Because \pifive{} was pretrained primarily on demonstrations from robots with parallel-jaw grippers \pifive{}~\cite{intelligence2025pi_}, its stronger performance on Trossen may reflect closer embodiment alignment. Its lower performance with dexterous hands suggests less effective transfer to that embodiment.

\subsection{Sample Efficiency}
\label{sec:sample_efficiency}

\begin{figure}[t]
    \centering
    \includegraphics[width=\columnwidth]{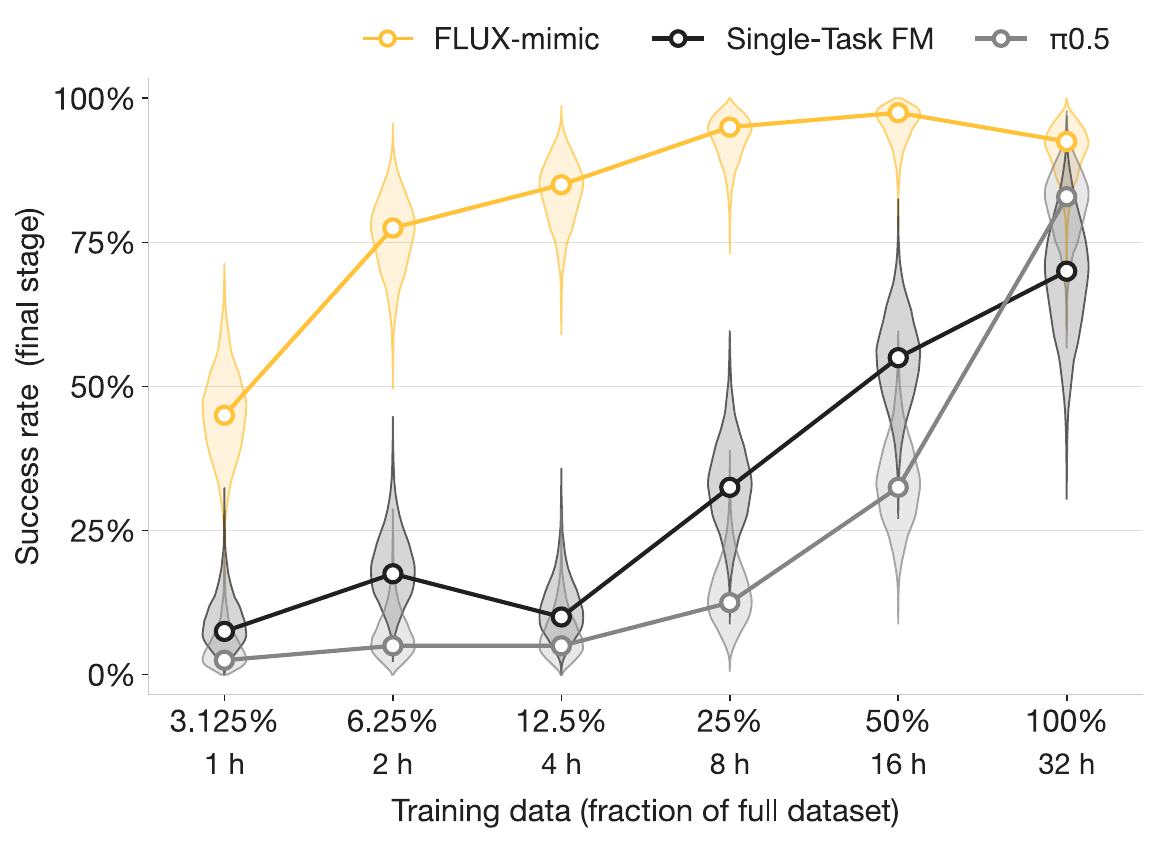}
    \caption{\textbf{Sample efficiency on soft part kitting with Franka.} Success rate (final stage) as a function of action-data budget, from 1\,h to 32\,h (3.125\% to 100\% of the full task dataset). Each checkpoint is evaluated over 40 rollouts. Markers show the empirical success rate at each data budget; shaded lobes show the posterior distribution over the true success rate.
    }
    \label{fig:sample_efficiency}
\end{figure}

To investigate how the amount of task data affects performance, we train all three methods on soft part kitting with the Franka platform using 1, 2, 4, 8, 16, and 32 hours of data.
In this experiment, all policies are trained in a single-task setting.
We use identical training subsets for the three methods at each budget.
The subsets are nested, such that each larger subset contains all examples from the smaller subsets.
The reported data budgets refer to policy training and exclude the backbone pretraining described in Section~\ref{sec:video_pretraining}.
We evaluate each resulting checkpoint over 40 episodes and report the full-task success rate.

Figure~\ref{fig:sample_efficiency} compares the methods at each data budget. The shaded lobes show the posterior distribution over the true success rate given $k$ successes in $n{=}40$ episodes, $\mathrm{Beta}(k{+}1,\,n{-}k{+}1)$ under a uniform prior. 
\model{} achieves the highest observed success rate across all six budgets, with the largest advantages in the low-data regime.
With only one hour of demonstrations, corresponding to 3.125\% of the full task dataset, \model{} reaches 45.0\% success, compared to 7.5\% for FM and 2.5\% for \pifive{}.

At eight hours, \model{} achieves 95.0\% success, exceeding the performance of both baselines trained on the full 32-hour dataset: 70.0\% for FM and 82.5\% for \pifive{}.
Thus, \model{} achieves higher success with one quarter of the action-labeled demonstrations.

The gap narrows at the full 32-hour budget, where \model{} achieves 92.5\% success, exceeding FM by 22.5 percentage points and \pifive{} by 10.0 percentage points.
Performance appears to plateau at the higher data budgets, with the variation between 16 and 32 hours consistent within the error bounds.
Overall, these results show that the pretrained VAM can achieve high task success with substantially less action data for policy training, while the baselines require larger datasets to approach its performance.

\subsection{Video Prediction Loss}
\label{sec:video_loss_ablation}

\begin{figure}[t]
    \centering
    \includegraphics[width=0.7\columnwidth]{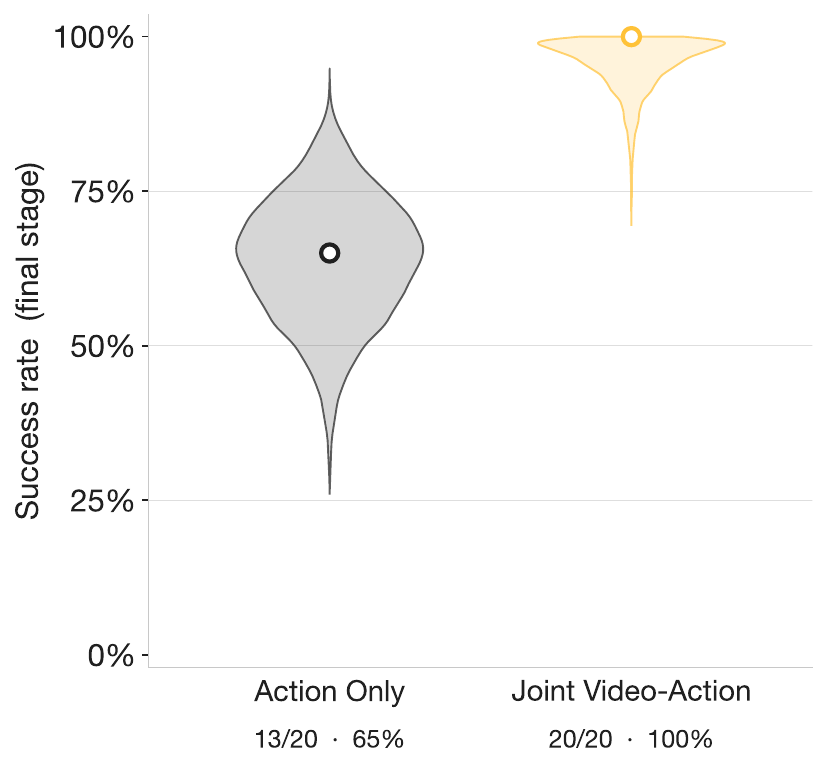}
    \caption{%
        \textbf{Effect of the video prediction loss.}
        Full-task success rate on soft part kitting for \model{} trained
        with the video prediction loss ($\lambda_{\mathrm{video}}=1$, Joint
        Video-Action) and without it ($\lambda_{\mathrm{video}}=0$, Action
        Only). Each variant is evaluated over 20 rollouts. Markers show the empirical success rate; shaded lobes show the posterior distribution over the true success rate.
    }
    \label{fig:video_loss_ablation}
\end{figure}

To ablate the contribution of video supervision during policy training, we compare \model{} with and without the video prediction loss on soft part kitting.
The joint variant uses $\lambda_{\mathrm{video}}=1$ in Eq.~\ref{eq:policy_training}, whereas the action-only variant uses $\lambda_{\mathrm{video}}=0$.
Both variants are multitask policies initialized from the same pretrained video backbone and trained on the full training dataset, with the backbone's weights updated in both cases.
We evaluate the variants on full-task success over 20 rollouts each, testing whether visual prediction as an auxiliary objective improves downstream control.
The violin distributions use the same uniform-prior construction as Section~\ref{sec:sample_efficiency}, with $n=20$.
As shown in Fig.~\ref{fig:video_loss_ablation}, joint video-action training achieves 100\% success (20/20), compared to 65\% (13/20) without the video loss, an improvement of 35 percentage points.
This result supports retaining video prediction as an auxiliary objective during policy training for this task.

\section{Conclusion}
\label{sec:conclusion}

We present \model{}, a video-action model for industrial manipulation that combines a backbone pretrained on large-scale web and industrial task data with joint video-action policy training.
Across two bimanual platforms, \model{} outperforms both baselines on all four Franka tasks and matches \pifive{}'s average success rate on Trossen.
A sample-efficiency experiment on soft part kitting with Franka shows higher performance for \model{} across data budgets from 1 to 32 hours, with the largest advantages in the low-data regime.
An ablation on soft part kitting further shows that including an auxiliary video prediction objective during training improves task performance.
These results establish a more general recipe for training models for industrial manipulation: a physical prior provided by a generalist video backbone, grounding through industrial task data during pretraining, and using only a few hours of action-labeled demonstrations for policy training.
This approach enables scaling to the long tail of industrial processes, as collecting task-specific data is the dominant cost of deploying learned policies in the factory today.

\balance
\bibliographystyle{IEEEtran}
\bibliography{root}
\end{document}